\documentclass{article} 
\usepackage{iclr2020_conference,times}

\usepackage{amsmath,amsfonts,bm}

\def\eqref#1{equation~\ref{#1}}

\def\1{\bm{1}}

\DeclareMathAlphabet{\mathsfit}{\encodingdefault}{\sfdefault}{m}{sl}
\SetMathAlphabet{\mathsfit}{bold}{\encodingdefault}{\sfdefault}{bx}{n}

\usepackage{graphicx}
\usepackage{hyperref}
\usepackage{url}

\title{Read, Highlight and Summarize: A Hierarchical Neural Semantic Encoder-based Approach}

\author{Rajeev Bhatt Ambati \\
Department of Electrical Engineering \\
Pennsylvania State University \\
State College, PA 16801, USA \\
\texttt{rajeev24811@gmail.com} \\
\And
Saptarashmi Bandyopadhyay \\
Department of Computer Science and Engineering \\
Pennsylvania State University \\
State College, PA 16801, USA \\
\texttt{sbandyo20@gmail.com} \\
\And
Prasenjit Mitra \\
College of Information Science and Technology \\
Pennsylvania State University \\
State College, PA 16801, USA \\
\texttt{pum10@psu.edu}
}

\iclrfinalcopy 
\begin{document}

\maketitle

\begin{abstract}
Traditional sequence-to-sequence (seq2seq) models and other variations of the attention-mechanism such as hierarchical attention have been applied to the text summarization problem. Though there is a hierarchy in the way humans use language by forming paragraphs from sentences and sentences from words, hierarchical models have usually not worked that much better than their traditional seq2seq counterparts. This effect is mainly because either the hierarchical attention mechanisms are too sparse using hard attention or noisy using soft attention. In this paper, we propose a method based on extracting the highlights of a document; a key concept that is conveyed in a few sentences. In a typical text summarization dataset consisting of documents that are 800 tokens in length (average), capturing long-term dependencies is very important, \textit{e.g.}, the last sentence can be grouped with the first sentence of a document to form a summary. 
LSTMs (Long Short-Term Memory) proved useful for machine translation. However, they often fail to capture long-term dependencies while modeling long sequences. To address these issues, we have adapted Neural Semantic Encoders (NSE) to text summarization, a class of memory-augmented neural networks by improving its functionalities and proposed a novel hierarchical NSE that outperforms similar previous models significantly. The quality of summarization was improved by augmenting linguistic factors, namely lemma, and Part-of-Speech (PoS) tags, to each word in the dataset for improved vocabulary coverage and generalization. The hierarchical NSE model on factored dataset outperformed the state-of-the-art by nearly 4 ROUGE points. We further designed and used the first GPU-based self-critical Reinforcement Learning model.
\end{abstract}

\section{Introduction}

When there are a very large number of documents that need to be read in limited time, we often resort to reading summaries instead of the whole document. Automatically generating (abstractive) summaries is a problem with various applications, e.g., automatic authoring~\citep{banerjee-mitra-2015-wikikreator}. We have developed automatic text summarization systems that condense large documents into short and readable summaries. It can be used for both single (\textit{e.g}., \citet{rush-etal-2015-neural}, \citet{see-etal-2017-get} and \citet{Nallapati-etal-2017}) and multi-document summarization (\textit{e.g}.,\citet{celikyilmaz-etal-2018}, \citet{Nallapati-etal-2017}, \citet{DBLP:conf/gldv/HenssMG15}).

Text summarization is broadly classified into two categories: extractive (\textit{e.g}., \citet{Nallapati-etal-2017} and \citep{narayan-etal-2018-ranking}) and abstractive summarization (\textit{e.g}., \citet{nallapati-etal-2016}, \citet{chopra-etal-2016} and \citet{chen-bansal-2018-fast}). Extractive approaches select sentences from a given document and groups them to form concise summaries.
By contrast, abstractive approaches generate human-readable summaries
that primarily capture the semantics of input documents and contain rephrased key content. The former task falls under the classification paradigm, and the latter belongs to the generative modeling paradigm, and therefore, it is a much harder problem to solve. The backbone of state-of-the-art summarization models is a typical encoder-decoder \citep{sutskever-etal-2014} architecture that has proved to be effective for various sequential modeling tasks such as machine translation, sentiment analysis, and natural language generation. It contains an encoder that maps the raw input word vector representations to a latent vector. Then, the decoder usually equipped with a variant of the attention mechanism \citep{bahdanau-etal-2014} uses the latent vectors to generate the output sequence, which is the summary in our case. These models are trained in a supervised learning setting where we minimize the cross-entropy loss between the predicted and the target summary. Encoder-decoder models have proved effective for short sequence tasks such as machine translation where the length of a sequence is less than 120 tokens. However, in text summarization, the length of the sequences vary from 400 to 800 tokens, and modeling long-term dependencies becomes increasingly difficult.

Despite the metric's known drawbacks, text summarization models are evaluated using ROUGE \citep{lin-2004-rouge}, a discrete similarity score between predicted and target summaries based on 1-gram, 2-gram, and n-gram overlap. Cross-entropy loss would be a convenient objective on which to train the model since ROUGE is not differentiable, but doing so would create a mismatch between metrics used for training and evaluation.
Though a particular summary scores well on ROUGE evaluation comparable to the target summary, it will be assigned lower probability by a supervised model. To tackle this problem, we have used a self-critic policy gradient method \citep{Rennie2016SelfCriticalST} to train the models directly using the ROUGE score as a reward. In this paper, we propose an architecture that addresses the issues discussed above.

\subsection{Problem Formulation}
\begin{figure}[htb]
    \centering
    \includegraphics[width=0.8\linewidth]{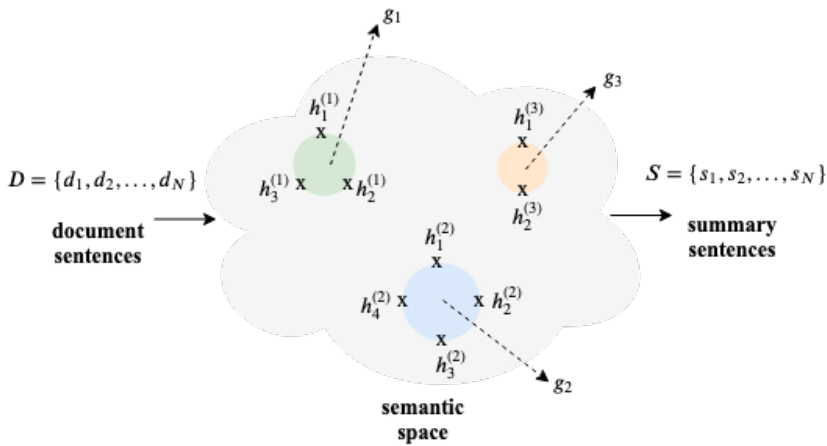}
    \caption{Document sentences are first projected into a semantic space typically by an encoder in a sequence-to-sequence model. $g_{1}, g_{2}, g_{3}$ are highlights of a document representing closely related sentence-semantics $\{h_{1}^{(1)}, h_{2}^{(1)}, h_{3}^{(1)}\}$, $\{h_{1}^{(2)}, h_{2}^{(2)}, h_{3}^{(2)}\}$, $\{h_{1}^{(3)}, h_{2}^{(3)}, h_{3}^{(3)}\}$ respectively. These highlights are then used by the decoder to form concise summaries.}
    \label{fig:model_flow}
\end{figure}

	Let $D=\{d_{1}, d_{2}, ..., d_{N}\}$ be the set of document sentences where each sentence $d_{i}, 1 \leq i \leq N$ is a set of words and $S=\{s_{1}, s_{2}, ..., s_{M}\}$ be the set of summary sentences. In general, most of the sentences in $D$ are a continuation of another sentence or related to each other, for example: in terms of factual details or pronouns used. So, dividing the document into multiple paragraphs as done by \citet{celikyilmaz-etal-2018} leaves out the possibility of a sentence-level dependency between the start and end of a document. Similarly, abstracting a single document sentence as done by \citet{chen-bansal-2018-fast} cannot include related information from multiple document sentences. In a good human-written summary, each summary sentence is a compressed version of a few document sentences. Mathematically,
    
    \begin{equation}
    	\forall s \in S, \exists d_{1}, d_{2}, ..., d_{K} \in D, \mid C(d_{1}, d_{2}, ..., d_{K}) = s
    \end{equation}
Where $C$ is a compressor we intend to learn. Figure \ref{fig:model_flow} represents the fundamental idea when using a sequence-to-sequence architecture. For a sentence $s$ in summary, the representations of all the related document sentences $d_{1}, d_{2}, ..., d_{K}$ are expected to form a cluster that represents a part of the highlight of the document.

First, we adapt the Neural Semantic Encoder (NSE) for text summarization by improving its attention mechanism and compose function. In a standard sequence-to-sequence model, the decoder has access to input sequence through hidden states of an LSTM \citep{Hochreiter:1997:LSM:1246443.1246450}, which suffers from the difficulties that we discussed
above. The NSE is equipped with an additional memory, which maintains
a rich representation of words by evolving over time.
We then propose a novel hierarchical NSE by using separate word memories for each sentence to enrich the word representations and a document memory to enrich the sentence representations, which performed better than its previous counterparts (\citet{nallapati-etal-2016}, \citet{Nallapati-etal-2017}, \citet{ling-rush-2017-coarse}). Finally, we use a maximum-entropy self-critic model to achieve better performance using ROUGE evaluation.

\section{Related Work}
 The first encoder-decoder for text summarziation is used by \citet{rush-etal-2015-neural} coupled with an attention mechanism. Though encoder-decoder models gave a state-of-the-art performance for Neural Machine Translation (NMT), the maximum sequence length used 
in NMT is just 100 tokens. Typical document lengths in text summarization vary from 400 to 800 tokens, and LSTM is not effective due to the loss in memory over time for very long sequences. \citet{nallapati-etal-2016} used hierarchical attention\citep{yang-etal-2016-hierarchical} to mitigate this effect where, a word LSTM is used to encode (decode) words, and a sentence LSTM is used to encode (decode) sentences. The use of two LSTMs separately for words and sentences improves the ability of the model to retain its memory for longer sequences. Additionally, \citet{nallapati-etal-2016} explored using a hierarchical model consisting of a feature-rich encoder incorporating position, Named Entity Recognition (NER) tag, Term Frequency (TF) and Inverse Document Frequency (IDF) scores. Since an RNN is a sequential model, computing at one time-step needs all of the previous time-steps to have computed before and is slow because the computation at all the time steps cannot be performed in parallel. \citet{chopra-etal-2016} used convolutional layers coupled with an attention mechanism \citep{bahdanau-etal-2014} to increase the speed of the encoder. Since the input to an RNN is fed sequentially, it is expected to capture the positional information. But both works \citet{nallapati-etal-2016} and \citet{chopra-etal-2016} found positional embeddings to be quite useful for reasons unknown. \citet{Nallapati-etal-2017} proposed an extractive summarization model that classifies sentences based on content, saliency, novelty, and position. To deal with out-of-vocabulary (OOV) words and to facilitate copying salient information from input sequence to the output, \citet{see-etal-2017-get} proposed a pointer-generator network that combines pointing \citep{Vinyals:2015:PN:2969442.2969540} with generation from vocabulary using a soft-switch. Attention models for longer sequences tend to be repetitive due to the decoder repeatedly attending to the same position from the encoder. To mitigate this issue, \citet{see-etal-2017-get} used a coverage mechanism to penalize a decoder from attending to same locations of an encoder. However, the pointer generator and the coverage model \citep{see-etal-2017-get} are still highly extractive; copying the whole article sentences 35\% of the time. \citet{paulus2018a} introduced an intra-attention model in which attention also depends on the predictions from previous time steps.

One of the main issues with sequence-to-sequence models is that optimization using the cross-entropy objective does not always provide excellent results because the models suffer from a mismatch between the training objective and the evaluation metrics such as ROUGE \citep{lin-2004-rouge} and METEOR \citep{banerjee-lavie-2005-meteor}. A popular algorithm to train a decoder is the teacher-forcing algorithm that minimizes the negative log-likelihood (cross-entropy loss) at each decoding time step given the previous ground-truth outputs. But during the testing stage, the prediction from the previous time-step is fed as input to the decoder instead of the ground truth. This exposure bias results in error accumulation over each time step because the model has never been exposed to its predictions during training. 
Instead, recent works show that summarization models can be trained 
using reinforcement learning (RL) where the ROUGE \citep{lin-2004-rouge} score is used as the reward (\citet{paulus2018a}, \citet{chen-bansal-2018-fast} and \citet{celikyilmaz-etal-2018}).

\citet{DBLP:conf/gldv/HenssMG15} made such an earlier attempt by using Q-learning for single-and multi-document summarization. Later, \citet{ling-rush-2017-coarse} proposed a coarse-to-fine hierarchical attention model to select a salient sentence using sentence attention using REINFORCE \citep{Williams:1992:SSG:139611.139614} and feed it to the decoder. \citet{narayan-etal-2018-ranking} used REINFORCE to rank sentences for extractive summarization. \citet{celikyilmaz-etal-2018} proposed deep communicating agents that operate over small chunks of a document, which is learned using a self-critical \citep{Rennie2016SelfCriticalST} training approach consisting of intermediate rewards. \citet{chen-bansal-2018-fast} used a advantage actor-critic (A2C) method to extract sentences followed by a decoder to form abstractive summaries. Our model does not suffer from their limiting assumption that a summary sentence is an abstracted version of a single source sentence. \citet{paulus2018a} trained their intra-attention model using a self-critical policy gradient algorithm \citep{Rennie2016SelfCriticalST}. Though an RL objective gives
a high ROUGE score, the output summaries are not readable by humans. To mitigate this problem, \citet{paulus2018a} used a weighted sum of supervised learning loss and RL loss.

Humans first form an abstractive representation of what they want to say and then try to put it into words while communicating. Though it seems intuitive that there is a hierarchy from sentence representation to words, as observed by both \citet{nallapati-etal-2016} and \citet{ling-rush-2017-coarse}, these hierarchical attention models failed to outperform a simple attention model \citep{rush-etal-2015-neural}. Unlike feedforward networks, RNNs are expected to capture the input sequence order. But strangely, positional embeddings are found to be effective (\citet{nallapati-etal-2016}, \citet{chopra-etal-2016}, \citet{ling-rush-2017-coarse} and \citet{Nallapati-etal-2017}). We explored a few approaches to solve these issues and improve the performance of neural models for abstractive summarization.

\section{Proposed Models}
In this section, we first describe the baseline Neural Semantic Encoder (NSE) class, discuss improvements to the compose function and attention mechanism, and then propose the Hierarchical NSE. Finally, we discuss the self-critic model that is used to boost the performance further using ROUGE evaluation.

\subsection{Neural Semantic Encoder:}
A Neural Semantic Encoder \citep{munkhdalai-yu-2017-neural-semantic} is a memory augmented neural network augmented with an encoding memory that supports read, compose, and write operations. Unlike the traditional sequence-to-sequence models, using an additional memory relieves the LSTM of the burden to remember the whole input sequence. Even compared to the attention-model \citep{bahdanau-etal-2014} which uses an additional context vector, the NSE has anytime access to the full input sequence through a much larger memory. The encoding memory is evolved using basic operations described as follows:

\begin{equation}\label{eq:1}
    o_{t} = f^{LSTM}_{read}(x_{t})
\end{equation}

\begin{equation}\label{eq:2}
    z_{t} = softmax(o^{T}_{t}M_{t-1})
\end{equation}

\begin{equation}\label{eq:3}
    m_{r,t} = z^{T}_{t}M_{t-1}
\end{equation}

\begin{equation}\label{eq:4}
    c_{t} = f^{MLP}_{c}(o_{t},m_{t,t})
\end{equation}

\begin{equation}\label{eq:5}
    h_{t} = f^{LSTM}_{w}(c_{t})
\end{equation}

\begin{equation}\label{eq:vanilla_update}
    M_{t} = M_{t-1}(\mathbf{1} - (z_{t} \otimes e_{k})^{T}) + (h_{t} \otimes e_{l})(z_{t} \otimes e_{k})^{T}
\end{equation}

Where, $x_{t} \in \mathbb{R}^D$ is the raw embedding vector at the current time-step. $f_{r}^{LSTM}$ , $f_{c}^{MLP}$ (Multi-Layer Perceptron), $f_{w}^{LSTM}$ be the read, compose and write operations respectively. $e_{l} \in R^{l}$ , $e_{k} \in R^{k}$ are vectors of ones, $\mathbf{1}$ is a matrix of ones and $\otimes$ is the outer product.

Instead of using the raw input, the read function $f_{r}^{LSTM}$ in equation \ref{eq:1} uses an LSTM to project the word embeddings to the internal space of memory $M_{t-1}$ to obtain the hidden states $o_{t}$. Now, the alignment scores $z_{t}$ of the past memory $M_{t-1}$ are calculated using $o_{t}$ as the key with a simple dot-product attention mechanism shown in equation \ref{eq:2}. A weighted sum gives the retrieved input memory that is used in equation \ref{eq:4} by a Multi-Layer Perceptron in composing new information. Equation \ref{eq:5} uses an LSTM and projects the composed states into the internal space of memory $M_{t-1}$ to obtain the write states $h_{t}$. Finally, in equation \ref{eq:vanilla_update}, the memory is updated by erasing the retrieved memory as per $z_{t}$ and writing as per the write vector $h_{t}$. This process is performed at each time-step throughout the input sequence. The encoded memories $\{M\}_{t=1}^{T}$ are similarly used by the decoder to obtain the write vectors $\{h\}_{t=1}^{T}$ that are eventually fed to projection and softmax layers to get the vocabulary distribution.

\subsection{Improved NSE}
Although the vanilla NSE described above performed well for machine translation, just a dot-product attention mechanism is too simplistic for text summarization. In machine translation, it is sufficient to compute the correlation between word-vectors from the semantic spaces of different languages. In contrast, text summarization also needs a word-sentence and sentence-sentence correlation along with the word-word correlation. So, in search of an attention mechanism with a better capacity to model the complex semantic relationships inherent in text summarization, we found that the additive attention mechanism \citep{bahdanau-etal-2014} given by the equation below performs well.
\begin{equation}
    z_{t} = softmax(v^{T}\tanh{(WM_{t-1} + Uo_{t} + b_{attn})})
    \label{eq:modified_attn}
\end{equation}

Where, $v, W, U, b_{attn}$ are learnable parameters. One other important difference is the compose function: a Multi-layer Perceptron (MLP) is enough for machine translation as the sequences are short in length. However, text summarization consists of longer sequences that have sentence-to-sentence dependencies, and a history of previously composed words is necessary for overcoming repetition \citep{rush-etal-2015-neural} and thereby maintaining novelty. A powerful function already at our disposal is the LSTM; we replaced the MLP with an LSTM, as shown below:

\begin{equation}
    h_{t} = f^{LSTM}_{w}(c_{t})
\end{equation}

In a standard text summarization task, due to the limited size of word vocabulary, out-of-vocabulary (OOV) words are replaced with [UNK] tokens. pointer-networks \citep{Vinyals:2015:PN:2969442.2969540} facilitate the ability to copy words from the input sequence to the output via pointing. Later, \citet{see-etal-2017-get} proposed a hybrid pointer-generator mechanism to improve upon pointing by retaining the ability to generate new words. It points to the words from the input sequence and generates new words from the vocabulary. A generation probability $p_{gen} \in (0, 1)$ is calculated using the retrieved memories, attention distribution, current input hidden state $o_{t}$ and write state $h_{t}$ as follows:

\begin{equation}
    p_{gen} = \sigma(W^{T}_{m}m_{r,t} + W^{T}_{h}h_{t} + W^{T}_{o}o_{t} + b_{ptr})
\end{equation}

Where, $W_{m}, W_{h}, W_{o}, b_{ptr}$ are learnable parameters, and $\sigma$ is the sigmoid activation function. Next, $p_{gen}$ is used as a soft switch to choose between generating a word from the vocabulary by sampling from $p_{vocab}$, or copying a word from the input sequence by sampling from the attention distribution $z_{t}$. For each document, we maintain an auxiliary vocabulary of OOV words in the input sequence. We obtain the following final probability distribution over the total extended vocabulary:

\begin{equation}
    p(w) = p_{gen}p_{vocab} + (1 - p_{gen})\sum_{i:w = w_{i}}z_{i}^{t}
\end{equation}

Note that if $w$ is an OOV word, then $p_{vocab}(w)$ is zero; similarly, if $w$ does not appear in the source document, then $\sum_{i:w = w_{i}} z_{i}^{t}$ is zero. The ability to produce OOV words is one of the primary advantages of the pointer-generator mechanism. We can also use a smaller vocabulary size and thereby speed up the computation of output projection and softmax layers.

\subsection{Hierarchical NSE}

\begin{figure*}[hbt!]
	\centering
    \includegraphics[width=0.8\linewidth]{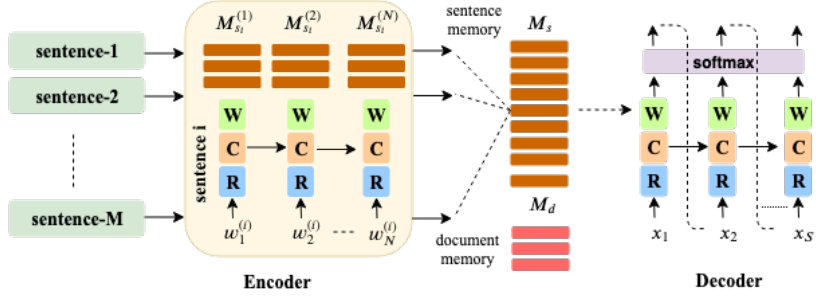}
    \caption{Hierarchical NSE: From a given article, all the $M$ sentences consisting of $N$ words each are processed by the NSE using read (R), compose (C) and write (W) operations. Each sentence memory is updated $N$ times by each word in the sentence ($\{M_{s_{i}}^{(k)}\}_{k=1}^{N}$). After the last encoder step, all the updated sentence memories $M_{s_{1}}^{N}, M_{s_{2}}^{N}, ..., M_{s_{M}}^{N}$ are concatenated to form the cumulative sentence memory $M_{s}$. The decoder then uses the cumulative sentence memory $M_{s}$ and document memory $M_{d}$ in a similar fashion to produce the write vectors $h_{t}$ that are passed through a softmax layer to obtain the vocabulary distribution.}
    \label{fig:hier_nse}
\end{figure*}

When humans read a document, we organize it in terms of word semantics followed by sentence semantics and then document semantics. In a text summarization task, after reading a document, sentences that have similar meanings or continual information are grouped together and then expressed in words. Such a hierarchical model was first introduced by \citet{yang-etal-2016-hierarchical} for document classification and later explored unsuccessfully for text summarization \citep{Nallapati-etal-2017}. In this work, we propose to use a hierarchical model with improved NSE to take advantage of both augmented memory and also the hierarchical document representation. We use a separate memory for each sentence to represent all the words of a sentence and a document memory to represent all sentences. Word memory composes novel words, and document memory composes novel sentences in the encoding process that can be later used to extract highlights and decode to summaries as shown in Figure \ref{fig:hier_nse}.

Let $D = \{(w_{ij})_{j=1}^{T_{in}}\}_{i=1}^{S_{in}}$ be the input document sequence, where $S_{in}$ is the number of sentences in a document and $T_{in}$ is the number of words per sentence. Let $\{M_{i}\}_{i=1}^{S_{in}}, M_{i} \in R^{T_{in} \times D}$ be the sentence memories that encode all the words in a sentence and $M^{d}, M^{d} \in R^{S_{in} \times D}$ be the document memory that encodes all the sentences present in the document. At each time-step, an input token $x_{t}$ is read and is used to retrieve aligned content from both corresponding sentence memory $M_{t}^{i, s}$ and document memory $M_{t}^{d}$. Please note that the retrieved document memory, which is a weighted combination of all the sentence representations forms a highlight. After composition, both the sentence and document memories are written simultaneously. This way, the words are encoded with contextual meaning, and also new simpler sentences are formed. The functionality of the model is as follows:
\begin{equation}
	o_{t} = f_{r}^{LSTM}(x_{t})
\end{equation}
\begin{equation}
    z_{t}^{s} = f_{attn}(M_{t-1}^{s}, o_{t})
\end{equation}
\begin{equation}
    z_{t}^{d} = f_{attn}(M_{t-1}^{d}, o_{t})
\end{equation}
\begin{equation}
    m_{r,t}^{s} = z_{t}^{s}M_{t-1}^{s}
\end{equation}
\begin{equation}
    m_{r,t}^{d} = z_{t}^{d}M_{t-1}^{d}
\end{equation}
\begin{equation}
    c_{t} = f_{c}^{LSTM}(Concat(o_{t}, m_{r,t}^{s}, m_{r,t}^{d}))
\end{equation}
\begin{equation}
    h_{t} = f_{w}^{LSTM}(c_{t})
\end{equation}
\begin{equation}
    M_{t}^{s} = Update(M_{t-1}^{s}, z_{t}^{s}, h_{t})
\end{equation}
\begin{equation}
    M_{t}^{d} = Update(M_{t-1}^{d}, z_{t}^{d}, h_{t})
\end{equation}
\begin{equation}
    M^{s} = 
    \begin{cases}
        M^{s_{i}}, 1 \leq i \leq S_{in}   &  \text{encoder-stage} \\
        Concat(\{M^{s_{i}}\}_{i=1}^{S_{in}})   & \text{decoder-stage} \\
    \end{cases}
\end{equation}

Where, $f_{attn}$ is the attention mechanism given by equation(\ref{eq:modified_attn}). $Update$ remains the same as the vanilla NSE given by equation(\ref{eq:vanilla_update})and $Concat$ is the vector concatenation. Please note that NSE \citep{munkhdalai-yu-2017-neural-semantic} has a concept of shared memory but we use multiple memories for representing words and a document memory for representing sentences, this is fundamentally different to a shared memory which does not have a concept of hierarchy.

\subsection{Self-Critical Sequence Training}

As discussed earlier, training in a supervised learning setting creates a mismatch between training and testing objectives. Also, feeding the ground-truth labels in training time-step creates an exposure bias while testing in which we feed the predictions from the previous time-step. Policy gradient methods overcome this by directly optimizing the non-differentiable metrics such as ROUGE \citep{lin-2004-rouge} and METEOR \citep{banerjee-lavie-2005-meteor}. It can be posed as a Markov Decision Process in which the set of actions $\mathcal{A}$ is the vocabulary and reward $\mathcal{R}$ is the ROUGE score itself. So, we should find a policy $\pi(\theta)$ such that the set of sampled words $\tilde{y} = \{\tilde{y}_{1}, \tilde{y}_{2}, ..., \tilde{y}_{T}\}$ achieves highest ROUGE score among all possible summaries.

We used the self-critical model of \citet{Rennie2016SelfCriticalST} proposed for image captioning. In self-critical sequence training, the REINFORCE algorithm \citep{Williams:1992:SSG:139611.139614} is used by modifying its baseline as the greedy output of the current model. At each time-step $t$, the model predicts two words: $\hat{y}_{t}$ sampled from $p(\hat{y}_{t} | \hat{y}_{1}, \hat{y}_{2}, ..., \hat{y}_{t-1}, x)$, the baseline output that is greedily generated by considering the most probable word from the vocabulary and $\tilde{y}_{t}$ sampled from the $p(\tilde{y}_{t} | \tilde{y}_{1}, \tilde{y}_{2}, ..., \tilde{y}_{t-1}, x)$. This model is trained using the following loss function:

\begin{equation}
	L_{rl} = (r(\tilde{y}) - r(\hat{y})) \sum_{t=1}^{T}-\log(p(\tilde{y}_{t} | \tilde{y}_{1}, \tilde{y}_{2}, ..., \tilde{y}_{t-1}, x))
\end{equation}
Using the above training objective, the model learns to generate samples with high probability and thereby increasing $r(\tilde{y})$ above $r(\hat{y})$. Additionally, we have used enthttps://stackoverflow.com/questions/19053077/looping-over-data-and-creating-individual-figuresropy regularization.
\begin{equation}
	\mathrm{H_{t}} = - \sum_{v=1}^{V}p(\tilde{y}_{t}=v)\log(p(\tilde{y}_{t}=v))
\end{equation}
\begin{equation}
	L = L_{rl} - \alpha \sum_{t=1}^{T}H_{t}
\end{equation}
Where, $p(\tilde{y}_{t})=p(\tilde{y}_{t} | \tilde{y}_{1}, \tilde{y}_{2}, ..., \tilde{y}_{t-1}, x)$ is the sampling probability and $V$ is the size of the vocabulary. It is similar to the exploration-exploitation trade-off. $\alpha$ is the regularization coefficient that explicitly controls this trade-off: a higher $\alpha$ corresponds to more exploration, and a lower $\alpha$ corresponds to more exploitation. We have found that all TensorFlow based open-source implementations of self-critic models use a function (\textbf{tf.py{\textunderscore}func}) that runs only on CPU and it is very slow. To the best of our knowledge, ours is the first GPU based implementation.

\section{Experiments and Results}

\subsection{Dataset}
We used the CNN/Daily Mail dataset \citep{nallapati-etal-2016}, which has been used as the standard benchmark to compare text summarization models. This corpus has 286,817 training pairs, 13,368 validation pairs, and 11,487 test pairs, as defined by their scripts. The source document in the training set has 766 words spanning 29.74 sentences on an average while the summaries consist of 53 words and 3.72 sentences \citep{nallapati-etal-2016}. The unique characteristics of this dataset such as long documents, and ordered multi-sentence summaries present exciting challenges, mainly because the proven sequence-to-sequence LSTM based models find it hard to learn long-term dependencies in long documents. We have used the same train/validation/test split and examples for a fair comparison with the existing models. 

The factoring of lemma and Part-of-Speech (PoS) tag of surface words, are observed \citep{bandyopadhyay-2019-factored} to increase the performance of NMT models in terms of BLEU score drastically. This is due to the improvement of the vocabulary coverage and better generalization. We have added a pre-processing step by incorporating the lemma and PoS tag to every word of the dataset and training the supervised model on the factored data. The process of extracting the lemma and the PoS tags has been described in \citet{bandyopadhyay-2019-factored}. Please refer to the appendix for an example of factoring.

\subsection{Training Settings}
For all the plain NSE models, we have truncated the article to a maximum of 400 tokens and the summary to 100 tokens. For the hierarchical NSE models, articles are truncated to have a maximum of 20 sentences and 20 words per sentence each. Shorter sequences are padded with `PAD` tokens. Since the factored models have lemma, PoS tag and the separator `\textbar{}` for each word, sequence lengths should be close to 3 times the non-factored counterparts. For practical reasons of memory and time, we have used 800 tokens per article and 300 tokens for the summary.

For all the models, including the pointer-generator model, we use a vocabulary size of 50,000 words for both source and target. Though some previous works \citep{nallapati-etal-2016} have used large vocabulary sizes of 150,000, since our models have a copy mechanism, smaller vocabulary is enough to obtain good performance. Large vocabularies increase the  computation time. Since memory plays a prominent role in retrieval and update, it is vital to start with a good initialization. We have used 300-dimensional pre-trained GloVe \citep{pennington-etal-2014-glove} word-vectors to represent the input sequence to a model. Sentence memories are initialized with GloVe word-vectors of all the words in that sentence. Document memories are initialized with vector representations of all the sentences where a sentence is represented with the average of the GloVe word-vectors of all its words. All the models are trained using the Adam optimizer with the default learning rate of 0.001. We have not applied any regularization as the usage of dropout, and $L_{2}$ penalty resulted in similar performance, however with a drastically increased training time.

The Hierarchical models process one sentence at a time, and hence attention distributions need less memory, and therefore, a larger batch size can be used, which in turn speeds up the training process. The non-factored model is trained on 7-NVIDIA Tesla-P100 GPUs with a batch size of 448 (64 examples per GPU); it takes approximately 45 minutes per epoch. Since the factored sequences are long, we used a batch size of 96 (12 examples per GPU) on 8-NVIDIA Tesla-V100 GPUs. The Hier model reaches optimal cross-entropy loss in just 8 epochs, unlike 33-35 epochs for both \citet{nallapati-etal-2016} and \citet{see-etal-2017-get}. For the self-critical model, training is started from the best supervised model with a learning rate of 0.00005 and manually changed to 0.00001 when needed with $\alpha=0.0001$ and the reported results are obtained after training for 15 days.

\begin{table*}[htb]
    \caption{ROUGE $F_{1}$ scores on the test set. Our hierarchical (Hier-NSE) model outperform previous hierarchical and pointer-generator models. Hier-NSE-factor is the factored model and Hier-NSE-SC is the self-critic model.}
    \label{tab:results_table}
    \begin{center}
        \begin{tabular}{|p{2cm}|p{6cm}|p{1cm}|p{1cm}|p{1cm}|}
        	\hline
            Paradigm & Models & \multicolumn{3}{| c |}{ROUGE (\% F-score)} \\
            \cline{3-5}
             & & 1 & 2 & L \\
            \hline
             & HierAttn \citep{nallapati-etal-2016} & 32.75 & 12.21 & 29.01 \\
             & abstractive model \citep{nallapati-etal-2016} & 35.46 & 13.30 & 32.65 \\
             & Pointer Generator \citep{see-etal-2017-get} & 36.44 & 15.66 & 33.42 \\
             Supervised Learning & Pointer Generator + coverage
             \citep{see-etal-2017-get} & 39.53 & 17.28 & 36.38 \\
             & Hier-NSE (ours) & 38.31 & 16.34 & 35.26 \\
             & Hier-NSE-factor (ours) & \textbf{45.58} & \textbf{26.81} & \textbf{41.17} \\
             \hline
             & MLE+RL, with intra-attention \citep{paulus2018a} & 39.87 & 15.82 & 36.90 \\
             Reinforcement Learning & DCA, MLE+RL
             \citep{celikyilmaz-etal-2018} & 41.69 & 19.47 & 37.92 \\
             & Hier-NSE-SC (ours) & 39.42 & 16.46 & \textbf{36.93} \\
            \hline
        \end{tabular}
    \end{center}
\end{table*}

\subsection{Evaluation}
All the models are evaluated using the standard metric ROUGE; we report the F1 scores for ROUGE-1, ROUGE-2, and ROUGE-L, which quantitively represent word-overlap, bigram-overlap, and longest common subsequence between reference summary and the summary that is to be evaluated. The results are obtained using \textit{pyrouge} package\footnote{https://pypi.org/project/pyrouge/}. The performance of various models and our improvements are summarized in Table \ref{tab:all_models_performance}. A direct implementation of NSE performed very poorly due to the simple dot-product attention mechanism. In NMT, a transformation from word-vectors in one language to another one (say English to French) using a mere matrix multiplication is enough because of the one-to-one correspondence between words and the underlying linear structure imposed in learning the word vectors \citep{pennington-etal-2014-glove}. However, in text summarization a word (sentence) could be a condensation of a group of words (sentences). Therefore, using a complex neural network-based attention mechanism proposed improved the performance. Both dot-product and additive \citep{bahdanau-etal-2014} mechanisms perform similarly for the NMT task, but the difference is more pronounced for the text summarization task simply because of the nature of the problem as described earlier. Replacing Multi-Layered Perceptron (MLP) in the NSE with an LSTM further improved the performance because it remembers what was previously composed and facilitates the composition of novel words. This also eliminates the need for additional mechanisms to penalize repetitions such as coverage \citep{see-etal-2017-get} and intra-attention \citep{paulus2018a}. Finally, using memories for each sentence enriches the corresponding word representation, and the document memory enriches the sentence representation that help the decoder. Please refer to the appendix for a few example outputs. Table \ref{tab:results_table} shows the results in comparison to the previous methods. Our hierarchical model outperforms \citet{nallapati-etal-2016} (HIER) by 5 ROUGE points. Our factored model achieves the new state-of-the-art (SoTA) result, outperforming \citet{celikyilmaz-etal-2018} by almost 4 ROUGE points.

\begin{table}[htb]
    \caption{Performance of various NSE models on CNN/Daily Mail corpus. Please note that the data is not factored here.}
    \label{tab:all_models_performance}
    \begin{center}
    \begin{tabular}{|c|c|c|c|}
        \hline
        Model & \multicolumn{3}{| c |}{ROUGE (\% F-score)} \\
        \cline{2-4}
         & 1 & 2 & L \\
        \hline
        Plain NSE & 7.99 & 0.86 & 7.52 \\
        NSE - improved attention & 25.47 & 8.96 & 24.01 \\
        NSE - improved compose & 30.86 & 11.42 & 29.04 \\
        Hierarchical NSE & 38.31 & 16.34 & 35.26 \\
        \hline
    \end{tabular}
    \end{center}
\end{table}

\section{Conclusion}
In this work, we presented a memory augmented neural network for the text summarization task that addresses the shortcomings of LSTM-based models. We applied a critical pre-processing step by factoring the dataset with inherent linguistic information that outperforms the state-of-the-art by a large margin. In the future, we will explore new sparse functions \citep{sparsemax} to enforce strict sparsity in selecting highlights out of sentences. The general framework of pre-processing, and extracting highlights can also be used with powerful pre-trained models like 
BERT \citep{bert} and XLNet \citep{xlnet}.

\bibliography{iclr2020_conference}

\begin{thebibliography}{27}
\providecommand{\natexlab}[1]{#1}
\providecommand{\url}[1]{\texttt{#1}}
\expandafter\ifx\csname urlstyle\endcsname\relax
  \providecommand{\doi}[1]{doi: #1}\else
  \providecommand{\doi}{doi: \begingroup \urlstyle{rm}\Url}\fi

\bibitem[Bahdanau et~al.(2014)Bahdanau, Cho, and Bengio]{bahdanau-etal-2014}
Dzmitry Bahdanau, Kyunghyun Cho, and Yoshua Bengio.
\newblock Neural machine translation by jointly learning to align and
  translate, 2014.
\newblock URL \url{http://arxiv.org/abs/1409.0473}.
\newblock cite arxiv:1409.0473Comment: Accepted at ICLR 2015 as oral
  presentation.

\bibitem[Bandyopadhyay(2019)]{bandyopadhyay-2019-factored}
Saptarashmi Bandyopadhyay.
\newblock Factored neural machine translation at {L}o{R}es{MT} 2019.
\newblock In \emph{Proceedings of the 2nd Workshop on Technologies for MT of
  Low Resource Languages}, pp.\  68--71, Dublin, Ireland, 20 August 2019.
  European Association for Machine Translation.
\newblock URL \url{https://www.aclweb.org/anthology/W19-6811}.

\bibitem[Banerjee \& Lavie(2005)Banerjee and Lavie]{banerjee-lavie-2005-meteor}
Satanjeev Banerjee and Alon Lavie.
\newblock {METEOR}: An automatic metric for {MT} evaluation with improved
  correlation with human judgments.
\newblock In \emph{Proceedings of the {ACL} Workshop on Intrinsic and Extrinsic
  Evaluation Measures for Machine Translation and/or Summarization}, pp.\
  65--72, Ann Arbor, Michigan, June 2005. Association for Computational
  Linguistics.
\newblock URL \url{https://www.aclweb.org/anthology/W05-0909}.

\bibitem[Banerjee \& Mitra(2015)Banerjee and
  Mitra]{banerjee-mitra-2015-wikikreator}
Siddhartha Banerjee and Prasenjit Mitra.
\newblock {W}iki{K}reator: Improving {W}ikipedia stubs automatically.
\newblock In \emph{Proceedings of the 53rd Annual Meeting of the Association
  for Computational Linguistics and the 7th International Joint Conference on
  Natural Language Processing (Volume 1: Long Papers)}, pp.\  867--877,
  Beijing, China, July 2015. Association for Computational Linguistics.
\newblock \doi{10.3115/v1/P15-1084}.
\newblock URL \url{https://www.aclweb.org/anthology/P15-1084}.

\bibitem[Celikyilmaz et~al.(2018)Celikyilmaz, Bosselut, He, and
  Choi]{celikyilmaz-etal-2018}
Asli Celikyilmaz, Antoine Bosselut, Xiaodong He, and Yejin Choi.
\newblock Deep communicating agents for abstractive summarization.
\newblock In \emph{Proceedings of the 2018 Conference of the North {A}merican
  Chapter of the Association for Computational Linguistics: Human Language
  Technologies, Volume 1 (Long Papers)}, pp.\  1662--1675, New Orleans,
  Louisiana, June 2018. Association for Computational Linguistics.
\newblock \doi{10.18653/v1/N18-1150}.
\newblock URL \url{https://www.aclweb.org/anthology/N18-1150}.

\bibitem[Chen \& Bansal(2018)Chen and Bansal]{chen-bansal-2018-fast}
Yen-Chun Chen and Mohit Bansal.
\newblock Fast abstractive summarization with reinforce-selected sentence
  rewriting.
\newblock In \emph{Proceedings of the 56th Annual Meeting of the Association
  for Computational Linguistics (Volume 1: Long Papers)}, pp.\  675--686,
  Melbourne, Australia, July 2018. Association for Computational Linguistics.
\newblock \doi{10.18653/v1/P18-1063}.
\newblock URL \url{https://www.aclweb.org/anthology/P18-1063}.

\bibitem[Chopra et~al.(2016)Chopra, Auli, and Rush]{chopra-etal-2016}
Sumit Chopra, Michael Auli, and Alexander~M. Rush.
\newblock Abstractive sentence summarization with attentive recurrent neural
  networks.
\newblock In \emph{Proceedings of the 2016 Conference of the North {A}merican
  Chapter of the Association for Computational Linguistics: Human Language
  Technologies}, pp.\  93--98, San Diego, California, June 2016. Association
  for Computational Linguistics.
\newblock \doi{10.18653/v1/N16-1012}.
\newblock URL \url{https://www.aclweb.org/anthology/N16-1012}.

\bibitem[Devlin et~al.(2018)Devlin, Chang, Lee, and Toutanova]{bert}
Jacob Devlin, Ming{-}Wei Chang, Kenton Lee, and Kristina Toutanova.
\newblock {BERT:} pre-training of deep bidirectional transformers for language
  understanding.
\newblock \emph{CoRR}, abs/1810.04805, 2018.
\newblock URL \url{http://arxiv.org/abs/1810.04805}.

\bibitem[Hen{\ss} et~al.(2015)Hen{\ss}, Mieskes, and
  Gurevych]{DBLP:conf/gldv/HenssMG15}
Stefan Hen{\ss}, Margot Mieskes, and Iryna Gurevych.
\newblock A reinforcement learning approach for adaptive single- and
  multi-document summarization.
\newblock In Bernhard Fisseni, Bernhard Schr{\"{o}}der, and Torsten Zesch
  (eds.), \emph{Proceedings of the International Conference of the German
  Society for Computational Linguistics and Language Technology, {GSCL} 2015,
  University of Duisburg-Essen, Germany, 30th September - 2nd October 2015},
  pp.\  3--12. {GSCL} e.V., 2015.
\newblock URL
  \url{http://gscl2015.inf.uni-due.de/wp-content/uploads/2016/02/GSCL-201503.pdf}.

\bibitem[Hochreiter \& Schmidhuber(1997)Hochreiter and
  Schmidhuber]{Hochreiter:1997:LSM:1246443.1246450}
Sepp Hochreiter and J\"{u}rgen Schmidhuber.
\newblock Long short-term memory.
\newblock \emph{Neural Comput.}, 9\penalty0 (8):\penalty0 1735--1780, November
  1997.
\newblock ISSN 0899-7667.
\newblock \doi{10.1162/neco.1997.9.8.1735}.
\newblock URL \url{http://dx.doi.org/10.1162/neco.1997.9.8.1735}.

\bibitem[Lin(2004)]{lin-2004-rouge}
Chin-Yew Lin.
\newblock {ROUGE}: A package for automatic evaluation of summaries.
\newblock In \emph{Text Summarization Branches Out}, pp.\  74--81, Barcelona,
  Spain, July 2004. Association for Computational Linguistics.
\newblock URL \url{https://www.aclweb.org/anthology/W04-1013}.

\bibitem[Ling \& Rush(2017)Ling and Rush]{ling-rush-2017-coarse}
Jeffrey Ling and Alexander Rush.
\newblock Coarse-to-fine attention models for document summarization.
\newblock In \emph{Proceedings of the Workshop on New Frontiers in
  Summarization}, pp.\  33--42, Copenhagen, Denmark, September 2017.
  Association for Computational Linguistics.
\newblock \doi{10.18653/v1/W17-4505}.
\newblock URL \url{https://www.aclweb.org/anthology/W17-4505}.

\bibitem[Martins \& Astudillo(2016)Martins and Astudillo]{sparsemax}
Andr{\'e} F.~T. Martins and Ram\'{o}n~F. Astudillo.
\newblock From softmax to sparsemax: A sparse model of attention and
  multi-label classification.
\newblock In \emph{Proceedings of the 33rd International Conference on
  International Conference on Machine Learning - Volume 48}, ICML'16, pp.\
  1614--1623. JMLR.org, 2016.
\newblock URL \url{http://dl.acm.org/citation.cfm?id=3045390.3045561}.

\bibitem[Munkhdalai \& Yu(2017)Munkhdalai and
  Yu]{munkhdalai-yu-2017-neural-semantic}
Tsendsuren Munkhdalai and Hong Yu.
\newblock Neural semantic encoders.
\newblock In \emph{Proceedings of the 15th Conference of the {E}uropean Chapter
  of the Association for Computational Linguistics: Volume 1, Long Papers},
  pp.\  397--407, Valencia, Spain, April 2017. Association for Computational
  Linguistics.
\newblock URL \url{https://www.aclweb.org/anthology/E17-1038}.

\bibitem[Nallapati et~al.(2016)Nallapati, Zhou, dos Santos,
  GuÌ‡l{\c{c}}ehre, and Xiang]{nallapati-etal-2016}
Ramesh Nallapati, Bowen Zhou, Cicero dos Santos, {\c{C}}a{\u{g}}lar
  GuÌ‡l{\c{c}}ehre, and Bing Xiang.
\newblock Abstractive text summarization using sequence-to-sequence {RNN}s and
  beyond.
\newblock In \emph{Proceedings of The 20th {SIGNLL} Conference on Computational
  Natural Language Learning}, pp.\  280--290, Berlin, Germany, August 2016.
  Association for Computational Linguistics.
\newblock \doi{10.18653/v1/K16-1028}.
\newblock URL \url{https://www.aclweb.org/anthology/K16-1028}.

\bibitem[Nallapati et~al.(2017)Nallapati, Zhai, and Zhou]{Nallapati-etal-2017}
Ramesh Nallapati, Feifei Zhai, and Bowen Zhou.
\newblock Summarunner: A recurrent neural network based sequence model for
  extractive summarization of documents.
\newblock In \emph{Proceedings of the Thirty-First AAAI Conference on
  Artificial Intelligence}, AAAI'17, pp.\  3075--3081. AAAI Press, 2017.
\newblock URL \url{http://dl.acm.org/citation.cfm?id=3298483.3298681}.

\bibitem[Narayan et~al.(2018)Narayan, Cohen, and
  Lapata]{narayan-etal-2018-ranking}
Shashi Narayan, Shay~B. Cohen, and Mirella Lapata.
\newblock Ranking sentences for extractive summarization with reinforcement
  learning.
\newblock In \emph{Proceedings of the 2018 Conference of the North {A}merican
  Chapter of the Association for Computational Linguistics: Human Language
  Technologies, Volume 1 (Long Papers)}, pp.\  1747--1759, New Orleans,
  Louisiana, June 2018. Association for Computational Linguistics.
\newblock \doi{10.18653/v1/N18-1158}.
\newblock URL \url{https://www.aclweb.org/anthology/N18-1158}.

\bibitem[Paulus et~al.(2018)Paulus, Xiong, and Socher]{paulus2018a}
Romain Paulus, Caiming Xiong, and Richard Socher.
\newblock A deep reinforced model for abstractive summarization.
\newblock In \emph{International Conference on Learning Representations}, 2018.
\newblock URL \url{https://openreview.net/forum?id=HkAClQgA-}.

\bibitem[Pennington et~al.(2014)Pennington, Socher, and
  Manning]{pennington-etal-2014-glove}
Jeffrey Pennington, Richard Socher, and Christopher Manning.
\newblock {G}love: Global vectors for word representation.
\newblock In \emph{Proceedings of the 2014 Conference on Empirical Methods in
  Natural Language Processing ({EMNLP})}, pp.\  1532--1543, Doha, Qatar,
  October 2014. Association for Computational Linguistics.
\newblock \doi{10.3115/v1/D14-1162}.
\newblock URL \url{https://www.aclweb.org/anthology/D14-1162}.

\bibitem[Rennie et~al.(2016)Rennie, Marcheret, Mroueh, Ross, and
  Goel]{Rennie2016SelfCriticalST}
Steven~J. Rennie, Etienne Marcheret, Youssef Mroueh, Jerret Ross, and Vaibhava
  Goel.
\newblock Self-critical sequence training for image captioning.
\newblock \emph{2017 IEEE Conference on Computer Vision and Pattern Recognition
  (CVPR)}, pp.\  1179--1195, 2016.

\bibitem[Rush et~al.(2015)Rush, Chopra, and Weston]{rush-etal-2015-neural}
Alexander~M. Rush, Sumit Chopra, and Jason Weston.
\newblock A neural attention model for abstractive sentence summarization.
\newblock In \emph{Proceedings of the 2015 Conference on Empirical Methods in
  Natural Language Processing}, pp.\  379--389, Lisbon, Portugal, September
  2015. Association for Computational Linguistics.
\newblock \doi{10.18653/v1/D15-1044}.
\newblock URL \url{https://www.aclweb.org/anthology/D15-1044}.

\bibitem[See et~al.(2017)See, Liu, and Manning]{see-etal-2017-get}
Abigail See, Peter~J. Liu, and Christopher~D. Manning.
\newblock Get to the point: Summarization with pointer-generator networks.
\newblock In \emph{Proceedings of the 55th Annual Meeting of the Association
  for Computational Linguistics (Volume 1: Long Papers)}, pp.\  1073--1083,
  Vancouver, Canada, July 2017. Association for Computational Linguistics.
\newblock \doi{10.18653/v1/P17-1099}.
\newblock URL \url{https://www.aclweb.org/anthology/P17-1099}.

\bibitem[Sutskever et~al.(2014)Sutskever, Vinyals, and Le]{sutskever-etal-2014}
Ilya Sutskever, Oriol Vinyals, and Quoc~V Le.
\newblock Sequence to sequence learning with neural networks.
\newblock In Z.~Ghahramani, M.~Welling, C.~Cortes, N.~D. Lawrence, and K.~Q.
  Weinberger (eds.), \emph{Advances in Neural Information Processing Systems
  27}, pp.\  3104--3112. Curran Associates, Inc., 2014.
\newblock URL
  \url{http://papers.nips.cc/paper/5346-sequence-to-sequence-learning-with-neural-networks.pdf}.

\bibitem[Vinyals et~al.(2015)Vinyals, Fortunato, and
  Jaitly]{Vinyals:2015:PN:2969442.2969540}
Oriol Vinyals, Meire Fortunato, and Navdeep Jaitly.
\newblock Pointer networks.
\newblock In \emph{Proceedings of the 28th International Conference on Neural
  Information Processing Systems - Volume 2}, NIPS'15, pp.\  2692--2700,
  Cambridge, MA, USA, 2015. MIT Press.
\newblock URL \url{http://dl.acm.org/citation.cfm?id=2969442.2969540}.

\bibitem[Williams(1992)]{Williams:1992:SSG:139611.139614}
Ronald~J. Williams.
\newblock Simple statistical gradient-following algorithms for connectionist
  reinforcement learning.
\newblock \emph{Mach. Learn.}, 8\penalty0 (3-4):\penalty0 229--256, May 1992.
\newblock ISSN 0885-6125.
\newblock \doi{10.1007/BF00992696}.
\newblock URL \url{https://doi.org/10.1007/BF00992696}.

\bibitem[Yang et~al.(2019)Yang, Dai, Yang, Carbonell, Salakhutdinov, and
  Le]{xlnet}
Zhilin Yang, Zihang Dai, Yiming Yang, Jaime~G. Carbonell, Ruslan Salakhutdinov,
  and Quoc~V. Le.
\newblock Xlnet: Generalized autoregressive pretraining for language
  understanding.
\newblock \emph{CoRR}, abs/1906.08237, 2019.
\newblock URL \url{http://arxiv.org/abs/1906.08237}.

\bibitem[Yang et~al.(2016)Yang, Yang, Dyer, He, Smola, and
  Hovy]{yang-etal-2016-hierarchical}
Zichao Yang, Diyi Yang, Chris Dyer, Xiaodong He, Alex Smola, and Eduard Hovy.
\newblock Hierarchical attention networks for document classification.
\newblock In \emph{Proceedings of the 2016 Conference of the North {A}merican
  Chapter of the Association for Computational Linguistics: Human Language
  Technologies}, pp.\  1480--1489, San Diego, California, June 2016.
  Association for Computational Linguistics.
\newblock \doi{10.18653/v1/N16-1174}.
\newblock URL \url{https://www.aclweb.org/anthology/N16-1174}.

\end{thebibliography}
\bibliographystyle{iclr2020_conference}

\appendix
\section{Appendix}
Figure \ref{fig:rl_sc} below shows the self-critical model. All the examples shown in Tables \ref{tab:output_ex1}-\ref{tab:output_ex4_fac} are chosen as per the shortest article lengths available due to space constraints.

\begin{figure}[htp]
	\centering
    \includegraphics[width=0.8\linewidth]{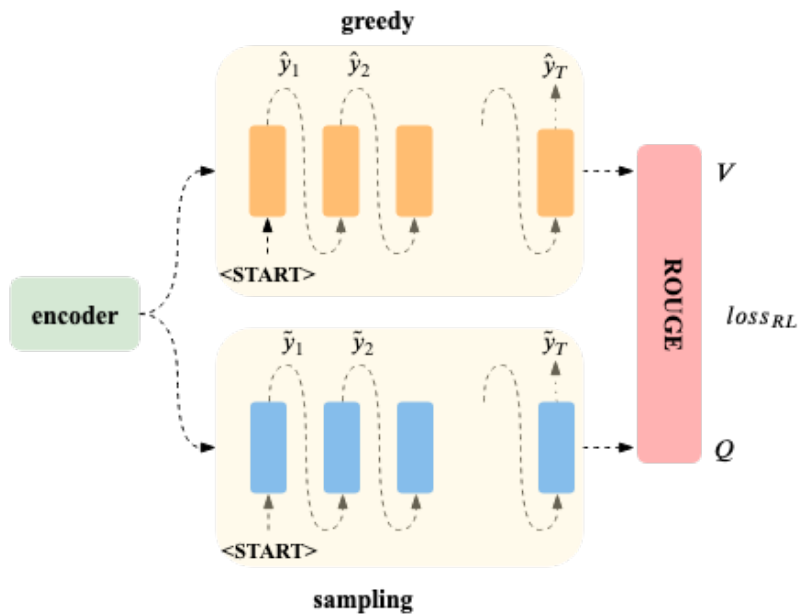}
    \caption{Self-Critic training reduces exposure bias and by learning a policy whose samples score better than the greedy samples that are used during test time in a supervised learning setting.}
    \label{fig:rl_sc}
\end{figure}

\begin{table}[htb]
    \caption{Sample outputs for both non-factored and factored input articles. While factoring, each surface word is augmented with lemma and PoS tag separated by \textbar{}.}
    \label{tab:output_ex1}
    \centering
    \begin{tabular}{| p{\linewidth} |}
        \hline 
          \textbf{Original Article} \\
         \hline
          {The build-up for the blockbuster fight between Floyd Mayweather and Manny Pacquiao in Las Vegas on May 2 steps up a gear on Tuesday night when the American holds an open workout for the media. The session will be streamed live across the world and you can watch it here from 12am UK.} \\
         \hline
          \textbf{Factored Article} \\
         \hline
          {The \textbar{} the \textbar{} DT build-up \textbar{} build-up \textbar{} NN for \textbar{} for \textbar{} IN the \textbar{} the \textbar{} DT blockbuster \textbar{} blockbust \textbar{} NN fight \textbar{} fight \textbar{} NN between \textbar{} between \textbar{} IN Floyd \textbar{} floyd \textbar{} NNP Mayweather \textbar{} mayweath \textbar{} NNP and \textbar{} and \textbar{} CC Manny \textbar{} manni \textbar{} NNP Pacquiao \textbar{} pacquiao \textbar{} NNP in \textbar{} in \textbar{} IN Las \textbar{} la \textbar{} NNP Vegas \textbar{} vega \textbar{} NNP on \textbar{} on \textbar{} IN May \textbar{} may \textbar{} NNP 2 \textbar{} 2 \textbar{} CD steps \textbar{} step \textbar{} NNS up \textbar{} up \textbar{} RB a \textbar{} a \textbar{} DT gear \textbar{} gear \textbar{} NN on \textbar{} on \textbar{} IN Tuesday \textbar{} tuesday \textbar{} NNP night \textbar{} night \textbar{} NN when \textbar{} when \textbar{} WRB the \textbar{} the \textbar{} DT American \textbar{} american \textbar{} NNP holds \textbar{} hold \textbar{} VBZ an \textbar{} an \textbar{} DT open \textbar{} open \textbar{} JJ workout \textbar{} workout \textbar{} NN for \textbar{} for \textbar{} IN the \textbar{} the \textbar{} DT media \textbar{} media \textbar{} NNS . \textbar{} . \textbar{} . The \textbar{} the \textbar{} DT session \textbar{} session \textbar{} NN will \textbar{} will \textbar{} MD be \textbar{} be \textbar{} VB streamed \textbar{} stream \textbar{} VBN live \textbar{} live \textbar{} JJ across \textbar{} across \textbar{} IN the \textbar{} the \textbar{} DT world \textbar{} world \textbar{} NN and \textbar{} and \textbar{} CC you \textbar{} you \textbar{} PRP can \textbar{} can \textbar{} MD watch \textbar{} watch \textbar{} VB it \textbar{} it \textbar{} PRP here \textbar{} here \textbar{} RB from \textbar{} from \textbar{} IN 12am \textbar{} 12am \textbar{} . \textbar{} .} \\
         \hline
          \textbf{GT Summary} \\
         \hline
          {floyd mayweather holds an open media workout from 12am uk -lrb- 7pm edt -rrb- . the american takes on manny pacquiao in las vegas on may 2 . mayweather 's training is being streamed live across the world .} \\
         \hline
          \textbf{Hier-NSE output} \\
         \hline
          {the build-up for the blockbuster fight between floyd mayweather and manny pacquiao in las vegas on may 2 steps up a gear on tuesday night . the session will be streamed live across the world and you can watch it here from uk . the session will be the media 's open workout for the media .} \\
         \hline
          \textbf{Hier-NSE-SC} \\
         \hline
          {the floyd mayweather and manny pacquiao in las vegas .
         the american holds an open workout for the media .
         will be streamed live across the world and .} \\
         \hline
          \textbf{GT summary (factored)} \\
         \hline
          {floyd \textbar{} floyd \textbar{} nnp mayweather \textbar{} mayweath \textbar{} nnp holds \textbar{} hold \textbar{} vbz an \textbar{} an \textbar{} dt open \textbar{} open \textbar{} jj media \textbar{} media \textbar{} nns workout \textbar{} workout \textbar{} nn from \textbar{} from \textbar{} in 12am \textbar{} 12am \textbar{} cd uk \textbar{} uk \textbar{} nnp -lrb- \textbar{} -lrb- \textbar{} vbd 7pm \textbar{} 7pm \textbar{} cd edt \textbar{} edt \textbar{} nnp -rrb- \textbar{} -rrb- \textbar{} nn . the \textbar{} the \textbar{} dt american \textbar{} american \textbar{} jj takes \textbar{} take \textbar{} vbz on \textbar{} on \textbar{} in manny \textbar{} manni \textbar{} nnp pacquiao \textbar{} pacquiao \textbar{} nnp in \textbar{} in \textbar{} in las \textbar{} la \textbar{} nnp vegas \textbar{} vega \textbar{} nnp on \textbar{} on \textbar{} in may \textbar{} may \textbar{} nnp 2 \textbar{} 2 \textbar{} cd . mayweather \textbar{} mayweath \textbar{} nnp 's \textbar{} 's \textbar{} pos training \textbar{} train \textbar{} nn is \textbar{} is \textbar{} vbz being \textbar{} be \textbar{} vbg streamed \textbar{} stream \textbar{} vbn live \textbar{} live \textbar{} jj across \textbar{} across \textbar{} in the \textbar{} the \textbar{} dt world \textbar{} world \textbar{} nn .} \\
         \hline
          \textbf{Hier-NSE output (factored)} \\
         \hline
          {the \textbar{} the \textbar{} dt session \textbar{} session \textbar{} nn will \textbar{} will \textbar{} md be \textbar{} be \textbar{} vb streamed \textbar{} stream \textbar{} vbn live \textbar{} live \textbar{} jj across \textbar{} across \textbar{} in the \textbar{} the \textbar{} dt world \textbar{} world \textbar{} nn and \textbar{} and \textbar{} cc you \textbar{} you \textbar{} prp can \textbar{} can \textbar{} md watch \textbar{} watch \textbar{} vb it \textbar{} it \textbar{} prp here \textbar{} here \textbar{} rb from \textbar{} from \textbar{} in 12am \textbar{} 12am \textbar{} nnp nnp \textbar{} nnp \textbar{} nnp . the \textbar{} the \textbar{} dt american \textbar{} american \textbar{} nnp holds \textbar{} hold \textbar{} vbz an \textbar{} an \textbar{} dt open \textbar{} open \textbar{} jj workout \textbar{} workout \textbar{} nn for \textbar{} for \textbar{} in the \textbar{} the \textbar{} dt media \textbar{} media \textbar{} nns .} \\
         \hline
    \end{tabular}
\end{table}

\begin{table}[htb]
    \caption{Sample outputs for both non-factored and factored input articles. While factoring, each surface word is augmented with lemma and PoS tag separated by \textbar{}.}
    \label{tab:output_ex2}
    \centering
    \begin{tabular}{| p{\linewidth} |}
        \hline
           \textbf{Original Article} \\
         \hline
          {-LRB- CNN -RRB- Justin Timberlake and Jessica Biel , welcome to parenthood . The celebrity couple announced the arrival of their son , Silas Randall Timberlake , in statements to People . `` Silas was the middle name of Timberlake 's maternal grandfather Bill Bomar , who died in 2012 , while Randall is the musician 's own middle name , as well as his father 's first , '' People reports . The couple announced the pregnancy in January , with an Instagram post . It is the first baby for both .} \\
         \hline
          \textbf{Factored Article} \\
         \hline
          {-LRB- \textbar{} -lrb- \textbar{} JJ CNN \textbar{} cnn \textbar{} NNP -RRB- \textbar{} -rrb- \textbar{} NNP Justin \textbar{} justin \textbar{} NNP Timberlake \textbar{} timberlak \textbar{} NNP and \textbar{} and \textbar{} CC Jessica \textbar{} jessica \textbar{} NNP Biel \textbar{} biel \textbar{} NNP , \textbar{} , \textbar{} , welcome \textbar{} welcom \textbar{} NN to \textbar{} to \textbar{} TO parenthood \textbar{} parenthood \textbar{} NN . \textbar{} . \textbar{} .
         The \textbar{} the \textbar{} DT celebrity \textbar{} celebr \textbar{} NN couple \textbar{} coupl \textbar{} NN announced \textbar{} announc \textbar{} VBD the \textbar{} the \textbar{} DT arrival \textbar{} arriv \textbar{} NN of \textbar{} of \textbar{} IN their \textbar{} their \textbar{} PRP son \textbar{} son \textbar{} NN , \textbar{} , \textbar{} , Silas \textbar{} sila \textbar{} NNP Randall \textbar{} randal \textbar{} NNP Timberlake \textbar{} timberlak \textbar{} NNP , \textbar{} , \textbar{} , in \textbar{} in \textbar{} IN statements \textbar{} statement \textbar{} NNS to \textbar{} to \textbar{} TO People \textbar{} peopl \textbar{} NNS . \textbar{} . \textbar{} . `` \textbar{} `` \textbar{} `` Silas \textbar{} sila \textbar{} NNP was \textbar{} wa \textbar{} VBD the \textbar{} the \textbar{} DT middle \textbar{} middl \textbar{} JJ name \textbar{} name \textbar{} NN of \textbar{} of \textbar{} IN Timberlake \textbar{} timberlak \textbar{} NNP 's \textbar{} 's \textbar{} POS maternal \textbar{} matern \textbar{} JJ grandfather \textbar{} grandfath \textbar{} NN Bill \textbar{} bill \textbar{} NNP Bomar \textbar{} bomar \textbar{} NNP , \textbar{} , \textbar{} , who \textbar{} who \textbar{} WP died \textbar{} die \textbar{} VBD in \textbar{} in \textbar{} IN 2012 \textbar{} 2012 \textbar{} CD , \textbar{} , \textbar{} , while \textbar{} while \textbar{} IN Randall \textbar{} randal \textbar{} NNP is \textbar{} is \textbar{} VBZ the \textbar{} the \textbar{} DT musician \textbar{} musician \textbar{} NN 's \textbar{} 's \textbar{} POS own \textbar{} own \textbar{} JJ middle \textbar{} middl \textbar{} NN name \textbar{} name \textbar{} NN , \textbar{} , \textbar{} , as \textbar{} as \textbar{} RB well \textbar{} well \textbar{} RB as \textbar{} as \textbar{} IN his \textbar{} hi \textbar{} PRP father \textbar{} father \textbar{} NN 's \textbar{} 's \textbar{} POS first \textbar{} first \textbar{} JJ , \textbar{} , \textbar{} , '' \textbar{} '' \textbar{} '' People \textbar{} peopl \textbar{} NNP reports \textbar{} report \textbar{} NNS . \textbar{} . \textbar{} . The \textbar{} the \textbar{} DT couple \textbar{} coupl \textbar{} NN announced \textbar{} announc \textbar{} VBD the \textbar{} the \textbar{} DT pregnancy \textbar{} pregnanc \textbar{} NN in \textbar{} in \textbar{} IN January \textbar{} januari \textbar{} NNP , \textbar{} , \textbar{} , with \textbar{} with \textbar{} IN an \textbar{} an \textbar{} DT Instagram \textbar{} instagram \textbar{} NNP post \textbar{} post \textbar{} NN . \textbar{} . \textbar{} . It \textbar{} It \textbar{} PRP is \textbar{} is \textbar{} VBZ the \textbar{} the \textbar{} DT first \textbar{} first \textbar{} JJ baby \textbar{} babi \textbar{} NN for \textbar{} for \textbar{} IN both \textbar{} both \textbar{} DT . | . \textbar{} .} \\
         \hline
          \textbf{GT Summary} \\
         \hline
          {timberlake and biel welcome son silas randall timberlake .
         the couple announced the pregnancy in january .} \\
         \hline
          \textbf{Hier-NSE Output} \\
         \hline
          {`` silas was the middle name of timberlake 's maternal
         grandfather bill bomar ' the couple announced the pregnancy in
         january , with an instagram post . it is the first baby for both  .} \\
         \hline
          \textbf{Hier-NSE-SC} \\
         \hline
          {justin timberlake and jessica biel the couple of their son , .
         silas randall timberlake , in . the first baby for both .} \\
         \hline
          \textbf{GT summary (factored)} \\
         \hline
          {timberlake \textbar{} timberlak \textbar{} nnp and \textbar{} and \textbar{} cc biel \textbar{} biel \textbar{} nnp welcome \textbar{} welcom \textbar{} vbp son \textbar{} son \textbar{} nn silas \textbar{} sila \textbar{} nnp randall \textbar{} randal \textbar{} nnp timberlake \textbar{} timberlak \textbar{} nnp . the \textbar{} the \textbar{} dt couple \textbar{} coupl \textbar{} nn announced \textbar{} announc \textbar{} vbd the \textbar{} the \textbar{} dt
         pregnancy \textbar{} pregnanc \textbar{} nn in \textbar{} in \textbar{} in january \textbar{} januari \textbar{} nnp .} \\
         \hline
          \textbf{Hier-NSE Output (factored)} \\
         \hline
          {justin \textbar{} justin \textbar{} nnp timberlake \textbar{} nnp \textbar{} nnp and \textbar{} and \textbar{} cc jessica \textbar{} jessica \textbar{} nnp nnp \textbar{} nnp \textbar{} nnp are \textbar{} are \textbar{} [UNK] in \textbar{} in \textbar{} in statements \textbar{} statement \textbar{} nns to \textbar{} to \textbar{} to people \textbar{} peopl \textbar{} nns . he \textbar{} he \textbar{} nnp is \textbar{} is \textbar{} vbz the \textbar{} the \textbar{} dt first \textbar{} first \textbar{} jj baby \textbar{} vbz \textbar{} nn for \textbar{} for \textbar{} in both \textbar{} both \textbar{} dt . timberlake \textbar{} [UNK] \textbar{} jj bill \textbar{} bill \textbar{} nn , \textbar{} , \textbar{} , the \textbar{} the \textbar{} dt couple \textbar{}  \textbar{} nn 's \textbar{} 's \textbar{} pos son \textbar{} son \textbar{} nn .} \\
         \hline
    \end{tabular}
\end{table}

\begin{table}[htb]
    \caption{Sample outputs from the hierarchical NSE and self-critical model.}
    \label{tab:output_ex3}
    \centering
    \begin{tabular}{| p{\linewidth} |}
        \hline
           \textbf{Original Article} \\
         \hline
          {-LRB- CNN -RRB- Once Hillary Clinton 's official announcement went online , social media responded in a big way , with terms like `` Hillary Clinton , '' '' \#Hillary2016 , '' and yes , even '' \#WhyImNotVotingforHillary '' trending . Certainly , you could n't go far on Twitter -LRB- even before Clinton tweeted her announcement -RRB- , without an opinion or thought on her new campaign -LRB- there were over 3 million views of her announcment tweets in one hour , and 750,000 Facebook video views so far by Sunday evening -RRB- . Some tweeted their immediate support , with one word :} \\
         \hline
          \textbf{GroundTruth Summary} \\
         \hline
          {response across social media led to multiple trending topics for hillary clinton 's presidential announcement . some responded to her video and her new campaign logo .} \\
         \hline
          \textbf{Hier-NSE Output} \\
         \hline
          {hillary clinton tweeted her announcement without an opinion or
         thought on her new campaign . some tweeted their immediate
         support , with one word : `` hillary clinton , '' yes .} \\
         \hline
          \textbf{Hier-NSE-SC} \\
         \hline
          {hillary clinton 's official announcement . clinton `` hillary clinton , '' . `` ' ' in the .} \\
         \hline
    \end{tabular}
\end{table}

\begin{table}[htb]
    \caption{Factored input and outputs for the same example used in Table \ref{tab:output_ex3}.}
    \label{tab:output_ex3_fac}
    \centering
    \begin{tabular}{| p{\linewidth} |}
        \hline
          \textbf{Article (Factored)} \\
         \hline        
          {-LRB- \textbar{} -lrb- \textbar{} JJ CNN \textbar{} cnn \textbar{} NNP -RRB- \textbar{} -rrb- \textbar{} NNP Once \textbar{} onc \textbar{} NNP Hillary \textbar{} hillari \textbar{} NNP Clinton \textbar{} clinton \textbar{} NNP 's \textbar{} 's \textbar{} POS official \textbar{} offici \textbar{} JJ announcement \textbar{} announc \textbar{} NN went \textbar{} went \textbar{} VBD online \textbar{} onlin \textbar{} NN , \textbar{} , \textbar{} , social \textbar{} social \textbar{} JJ media \textbar{} media \textbar{} NNS responded \textbar{} respond \textbar{} VBD in \textbar{} in \textbar{} IN a \textbar{} a \textbar{} DT big \textbar{} big \textbar{} JJ way \textbar{} way \textbar{} NN , \textbar{} , \textbar{} , with \textbar{} with \textbar{} IN terms \textbar{} term \textbar{} NNS like \textbar{} like \textbar{} IN `` \textbar{} `` \textbar{} `` Hillary \textbar{} hillari \textbar{} NNP Clinton \textbar{} clinton \textbar{} NNP , \textbar{} , \textbar{} , '' \textbar{} '' \textbar{} '' '' \textbar{} '' \textbar{} '' \# \textbar{} \# \textbar{} \# Hillary2016 \textbar{} hillary2016 \textbar{} NNP , \textbar{} , \textbar{} , '' \textbar{} '' \textbar{} '' and \textbar{} and \textbar{} CC yes \textbar{} ye \textbar{} UH , \textbar{} , \textbar{} , even \textbar{} even \textbar{} RB '' \textbar{} '' \textbar{} '' \# \textbar{} \# \textbar{} \# WhyImNotVotingforHillary \textbar{} whyimnotvotingforhillari \textbar{} NNP '' \textbar{} '' \textbar{} '' trending \textbar{} trend \textbar{} NN . \textbar{} . \textbar{} . Certainly \textbar{} certainli \textbar{} RB , \textbar{} , \textbar{} , you \textbar{} you \textbar{} PRP could \textbar{} could \textbar{} MD n't \textbar{} n't \textbar{} RB go \textbar{} go \textbar{} VB far \textbar{} far \textbar{} RB on \textbar{} on \textbar{} IN Twitter \textbar{} twitter \textbar{} NNP -LRB- \textbar{} -lrb- \textbar{} NNP even \textbar{} even \textbar{} RB before \textbar{} befor \textbar{} IN Clinton \textbar{} clinton \textbar{} NNP tweeted \textbar{} tweet \textbar{} VBD her \textbar{} her \textbar{} PRP announcement \textbar{} announc \textbar{} NN -RRB- \textbar{} -rrb- \textbar{} NN , \textbar{} , \textbar{} , without \textbar{} without \textbar{} IN an \textbar{} an \textbar{} DT opinion \textbar{} opinion \textbar{} NN or \textbar{} or \textbar{} CC thought \textbar{} thought \textbar{} NN on \textbar{} on \textbar{} IN her \textbar{} her \textbar{} PRP new \textbar{} new \textbar{} JJ campaign \textbar{} campaign \textbar{} NN -LRB- \textbar{} -lrb- \textbar{} NN there \textbar{} there \textbar{} EX were \textbar{} were \textbar{} VBD over \textbar{} over \textbar{} IN 3 \textbar{} 3 \textbar{} CD million \textbar{} million \textbar{} CD views \textbar{} view \textbar{} NNS of \textbar{} of \textbar{} IN her \textbar{} her \textbar{} PRP announcment \textbar{} announc \textbar{} JJ tweets \textbar{} tweet \textbar{} NNS in \textbar{} in \textbar{} IN one \textbar{} one \textbar{} CD hour \textbar{} hour \textbar{} NN , \textbar{} , \textbar{} , and \textbar{} and \textbar{} CC 750,000 \textbar{} 750,000 \textbar{} CD Facebook \textbar{} facebook \textbar{} NNP video \textbar{} video \textbar{} NN views \textbar{} view \textbar{} NNS so \textbar{} so \textbar{} RB far \textbar{} far \textbar{} RB by \textbar{} by \textbar{} IN Sunday \textbar{} sunday \textbar{} NNP evening \textbar{} even \textbar{} VBG -RRB- \textbar{} -rrb- \textbar{} NN . \textbar{} . \textbar{} . Some \textbar{} some \textbar{} DT tweeted \textbar{} tweet \textbar{} VBD their \textbar{} their \textbar{} PRP immediate \textbar{} immedi \textbar{} JJ support \textbar{} support \textbar{} NN , \textbar{} , \textbar{} , with \textbar{} with \textbar{} IN one \textbar{} one \textbar{} CD word \textbar{} word \textbar{} NN : \textbar{} : \textbar{} :} \\
         \hline
          \textbf{GT summary (factored)} \\
         \hline
          {response \textbar{} respons \textbar{} nnp across \textbar{} across \textbar{} in social \textbar{} social \textbar{} jj media \textbar{} media \textbar{} nns led \textbar{} led \textbar{} vbd to \textbar{} to \textbar{} to multiple \textbar{} multipl \textbar{} vb trending \textbar{} trend \textbar{} vbg topics \textbar{} topic \textbar{} nns for \textbar{} for \textbar{} in hillary \textbar{} hillari \textbar{} nnp clinton \textbar{} clinton \textbar{} nnp 's \textbar{} 's \textbar{} pos presidential \textbar{} presidenti \textbar{} jj announcement \textbar{} announc \textbar{} nn . some \textbar{} some \textbar{} dt responded \textbar{} respond \textbar{} vbd to \textbar{} to \textbar{} to her \textbar{} her \textbar{} prp video \textbar{} video \textbar{} nn and \textbar{} and \textbar{} cc her \textbar{} her \textbar{} prp new \textbar{}
         new \textbar{} jj campaign \textbar{} campaign \textbar{} nn logo \textbar{} logo \textbar{} nn .} \\
         \hline
          \textbf{Hier-NSE Output (factored)} \\
         \hline
          {hillary \textbar{} nnp \textbar{} nnp clinton \textbar{} clinton \textbar{} nnp 's \textbar{} 's \textbar{} pos official \textbar{} [UNK] \textbar{} jj announcement \textbar{} announc \textbar{} nn went \textbar{} went \textbar{} vbd online \textbar{} onlin \textbar{} nn . clinton \textbar{} clinton \textbar{} nnp tweeted \textbar{} tweet \textbar{} vbd her \textbar{} her \textbar{} prp new \textbar{} new \textbar{} jj campaign \textbar{} campaign \textbar{} nn , \textbar{} , \textbar{} , without \textbar{} without \textbar{} in an \textbar{} an \textbar{} dt opinion \textbar{} opinion \textbar{} nn or \textbar{} or \textbar{} cc thought \textbar{} thought \textbar{} vbd on \textbar{} on \textbar{} in twitter \textbar{} twitter \textbar{} nn , \textbar{} , \textbar{} , with \textbar{} with \textbar{} in terms \textbar{} term \textbar{} nns like \textbar{} like \textbar{} in `` \textbar{} `` \textbar{} `` hillary \textbar{} nnp \textbar{} nnp clinton \textbar{} clinton \textbar{} nnp , \textbar{} , \textbar{} , '' \textbar{} '' \textbar{} '' \# \textbar{} \# \textbar{} \# nnp \textbar{} nnp \textbar{} jj .} \\
         \hline
    \end{tabular}
\end{table}

\begin{table}[htb]
    \caption{Sample outputs from the hierarchical NSE and self-critical model.}
    \label{tab:output_ex4}
    \centering
    \begin{tabular}{| p{\linewidth} |}
        \hline
           \textbf{Original Article} \\
         \hline
          {Blackpool are in talks to sign Austria defender Thomas Piermayr . The 25-year-old has been training with the Championship club this week and they are keen to get him on board for what is expected to be confirmed as a campaign in League One next season . Piermayr is a free agent and had been playing for Colorado Rapids . The former Austria U21 international had a spell with Inverness Caledonian Thistle in 2011 . Thomas Piermayr -LRB- left , in action for the Colorado Rapids -RRB- tries to tackle Obafemi Martins last year} \\
         \hline
          \textbf{GroundTruth Summary} \\
         \hline
          {thomas piermayr has been training with blackpool this week . austrian defender is a free agent after leaving mls side colorado rapids . blackpool are bottom of the championship and look set to be relegated .} \\
         \hline
          \textbf{Hier-NSE Output} \\
         \hline
          {thomas  has been training with the championship club this week . the former austria u21 international had a spell with inverness caledonian thistle . blackpool are in talks to sign austria defender thomas  .} \\
         \hline
          \textbf{Hier-NSE-SC} \\
         \hline
          {blackpool are in talks to sign austria defender thomas  . has been training with the championship club this week . is a free agent and .} \\
         \hline
    \end{tabular}
\end{table}

\begin{table}[htb]
    \caption{Factored input and outputs for the same example used in Table \ref{tab:output_ex4}.}
    \label{tab:output_ex4_fac}
    \centering
    \begin{tabular}{| p{\linewidth} |}     
        \hline
          \textbf{Factored Article} \\
         \hline
          {Blackpool \textbar{} blackpool \textbar{} NNP are \textbar{} are \textbar{} VBP in \textbar{} in \textbar{} IN talks \textbar{} talk \textbar{} NNS to \textbar{} to \textbar{} TO sign \textbar{} sign \textbar{} VB Austria \textbar{} austria \textbar{} NNP defender \textbar{} defend \textbar{} NN Thomas \textbar{} thoma \textbar{} NNP Piermayr \textbar{} piermayr \textbar{} NNP . \textbar{} . \textbar{} . The \textbar{} the \textbar{} DT 25-year-old \textbar{} 25-year-old \textbar{} JJ has \textbar{} ha \textbar{} VBZ been \textbar{} been \textbar{} VBN training \textbar{} train \textbar{} VBG with \textbar{} with \textbar{} IN the \textbar{} the \textbar{} DT Championship \textbar{} championship \textbar{} NNP club \textbar{} club \textbar{} NN this \textbar{} thi \textbar{} DT week \textbar{} week \textbar{} NN and \textbar{} and \textbar{} CC they \textbar{} they \textbar{} PRP are \textbar{} are \textbar{} VBP keen \textbar{} keen \textbar{} JJ to \textbar{} to \textbar{} TO get \textbar{} get \textbar{} VB him \textbar{} him \textbar{} PRP on \textbar{} on \textbar{} IN board \textbar{} board \textbar{} NN for \textbar{} for \textbar{} IN what \textbar{} what \textbar{} WP is \textbar{} is \textbar{} VBZ expected \textbar{} expect \textbar{} VBN to \textbar{} to \textbar{} TO be \textbar{} be \textbar{} VB confirmed \textbar{} confirm \textbar{} VBN as \textbar{} as \textbar{} IN a \textbar{} a \textbar{} DT campaign \textbar{} campaign \textbar{} NN in \textbar{} in \textbar{} IN League \textbar{} leagu \textbar{} NNP One \textbar{} one \textbar{} NNP next \textbar{} next \textbar{} JJ season \textbar{} season \textbar{} NN . \textbar{} . \textbar{} . Piermayr \textbar{} piermayr \textbar{} NNP is \textbar{} is \textbar{} VBZ a \textbar{} a \textbar{} DT free \textbar{} free \textbar{} JJ agent \textbar{} agent \textbar{} NN and \textbar{} and \textbar{} CC had \textbar{} had \textbar{} VBD been \textbar{} been \textbar{} VBN playing \textbar{} play \textbar{} VBG for \textbar{} for \textbar{} IN Colorado \textbar{} colorado \textbar{} NNP Rapids \textbar{} rapid \textbar{} NNP . \textbar{} . \textbar{} . The \textbar{} the \textbar{} DT former \textbar{} former \textbar{} JJ Austria \textbar{} austria \textbar{} NNP U21 \textbar{} u21 \textbar{} NNP international \textbar{} intern \textbar{} JJ had \textbar{} had \textbar{} VBD a \textbar{} a \textbar{} DT spell \textbar{} spell \textbar{} NN with \textbar{} with \textbar{} IN Inverness \textbar{} inver \textbar{} NNP Caledonian \textbar{} caledonian \textbar{} NNP Thistle \textbar{} thistl \textbar{} NNP in \textbar{} in \textbar{} IN 2011 \textbar{} 2011 \textbar{} CD . \textbar{} . \textbar{} . Thomas \textbar{} thoma \textbar{} NNP Piermayr \textbar{} piermayr \textbar{} NNP -LRB- \textbar{} -lrb- \textbar{} NNP left \textbar{} left \textbar{} VBD , \textbar{} , \textbar{} , in \textbar{} in \textbar{} IN action \textbar{} action \textbar{} NN for \textbar{} for \textbar{} IN the \textbar{} the \textbar{} DT Colorado \textbar{} colorado \textbar{} NNP Rapids \textbar{} rapid \textbar{} NNP -RRB- \textbar{} -rrb- \textbar{} NNP tries \textbar{} tri \textbar{} VBZ to \textbar{} to \textbar{} TO tackle \textbar{} tackl \textbar{} VB Obafemi \textbar{} obafemi \textbar{} NNP Martins \textbar{} martin \textbar{} NNP last \textbar{} last \textbar{} JJ year \textbar{} year \textbar{} NN} \\
         \hline         
          \textbf{GT summary (factored)} \\
         \hline
          {thomas \textbar{} thoma \textbar{} nnp piermayr \textbar{} piermayr \textbar{} nnp has \textbar{} ha \textbar{} vbz been \textbar{} been \textbar{} vbn training \textbar{} train \textbar{} vbg with \textbar{} with \textbar{} in blackpool \textbar{} blackpool \textbar{} nnp this \textbar{} thi \textbar{} dt week \textbar{} week \textbar{} nn .
         austrian \textbar{} austrian \textbar{} jj defender \textbar{} defend \textbar{} nn is \textbar{} is \textbar{} vbz a \textbar{} a \textbar{} dt free \textbar{} free \textbar{} jj agent \textbar{} agent \textbar{} nn after \textbar{} after \textbar{} in leaving \textbar{} leav \textbar{} vbg mls \textbar{} ml \textbar{} nnp side \textbar{} side \textbar{} nn colorado \textbar{} colorado \textbar{} nnp rapids \textbar{} rapid \textbar{} nnp . blackpool \textbar{} blackpool \textbar{} nnp are \textbar{} are \textbar{} vbp bottom \textbar{} bottom \textbar{} nn of \textbar{} of \textbar{} in the \textbar{} the \textbar{} dt championship \textbar{} championship \textbar{} nnp and \textbar{} and \textbar{} cc look \textbar{} look \textbar{} vb set \textbar{} set \textbar{} vbn to \textbar{} to \textbar{} to be \textbar{} be \textbar{} vb relegated \textbar{} releg \textbar{} vbn .} \\
         \hline
          \textbf{Hier-NSE Output (factored)} \\
         \hline
          {the \textbar{} the \textbar{} dt 25-year-old \textbar{} 25-year-old \textbar{} jj has \textbar{} ha \textbar{} vbz been \textbar{} been \textbar{} vbn training \textbar{} train \textbar{} nnp with \textbar{} with \textbar{} in the \textbar{} the \textbar{} dt championship \textbar{} championship \textbar{} nnp club \textbar{} club \textbar{} nnp this \textbar{} thi \textbar{} dt week \textbar{} week \textbar{} nn . the \textbar{} the \textbar{} dt former \textbar{} former \textbar{} jj austria \textbar{} austria \textbar{} nnp u21 \textbar{} u21 \textbar{} nnp international \textbar{} intern \textbar{} jj had \textbar{} had \textbar{} vbd a \textbar{} a \textbar{} dt spell \textbar{} spell \textbar{} nn with \textbar{} with \textbar{} in nnp \textbar{} nnp \textbar{} nnp nnp \textbar{} nnp \textbar{} nnp .} \\
         \hline
    \end{tabular}
\end{table}

\end{document}